\documentclass[10pt,twocolumn,letterpaper]{article}

\usepackage{cvpr}      

\usepackage{graphicx}
\usepackage{amsmath}
\usepackage{amssymb}
\usepackage{booktabs}
\usepackage[accsupp]{axessibility}

\usepackage[pagebackref,breaklinks,colorlinks]{hyperref}

\usepackage[capitalize]{cleveref}
\crefname{section}{Sec.}{Secs.}
\Crefname{section}{Section}{Sections}
\Crefname{table}{Table}{Tables}
\crefname{table}{Tab.}{Tabs.}

\usepackage{algorithm}
\usepackage{algorithmic}
\usepackage{multirow}
\usepackage{bm} 

\usepackage{subfloat}
\def\eg{\emph{e.g.}}
\def\ie{\emph{i.e.}}
\def\etc{\emph{etc}}

\def\cvprPaperID{1680} 
\def\confName{CVPR}
\def\confYear{2022}

\begin{document}

\title{Defensive Patches for Robust Recognition in the Physical World}

\author{Jiakai Wang\textsuperscript{\dag}, Zixin Yin\textsuperscript{\dag}, Pengfei Hu\textsuperscript{\dag}, Aishan Liu\textsuperscript{\dag},\\ Renshuai Tao\textsuperscript{\dag},
Haotong Qin\textsuperscript{\dag},Xianglong Liu\textsuperscript{\dag}\textsuperscript{\thanks{Corresponding author}},Dacheng Tao\textsuperscript{\ddag}\\
\fontsize{11.0pt}{\baselineskip}\selectfont 
\textsuperscript{\dag}State Key Lab of Software Development Environment, Beihang University, Beijing, China\\
\textsuperscript{\ddag}JD Explore Academy, Beijing, China\\

{\tt\small \{jk\_buaa\_scse, yzx835, iamparasite, liuaishan\}@buaa.edu.cn}\\
{\tt\small \{rstao, qinhaotong,xlliu\}@buaa.edu.cn},
{\tt\small dacheng.tao@gmail.com}} 

\maketitle

\begin{abstract}

To operate in real-world high-stakes environments, deep learning systems have to endure noises that have been continuously thwarting their robustness. Data-end defense, which improves robustness by operations on input data instead of modifying models, has attracted intensive attention due to its feasibility in practice. However, previous data-end defenses show low generalization against diverse noises and weak transferability across multiple models. Motivated by the fact that robust recognition depends on both local and global features, we propose a defensive patch generation framework to address these problems by helping models better exploit these features. For the generalization against diverse noises, we inject class-specific identifiable patterns into a confined local patch prior, so that defensive patches could preserve more recognizable features towards specific classes, leading models for better recognition under noises. For the transferability across multiple models, we guide the defensive patches to capture more global feature correlations within a class, so that they could activate model-shared global perceptions and transfer better among models. Our defensive patches show great potentials to improve 
application robustness in practice by simply sticking them around target objects. Extensive experiments show that we outperform others by large margins (improve 20+\% accuracy for both adversarial and corruption robustness on average in the digital and physical world).\footnote{Our codes are available at \url{https://github.com/nlsde-safety-team/DefensivePatch}.}

\end{abstract}

\section{Introduction}
\label{sec:intro}

\begin{figure}
    \centering
    \centering
    \subfloat[]{
    \includegraphics[width=0.45\linewidth]{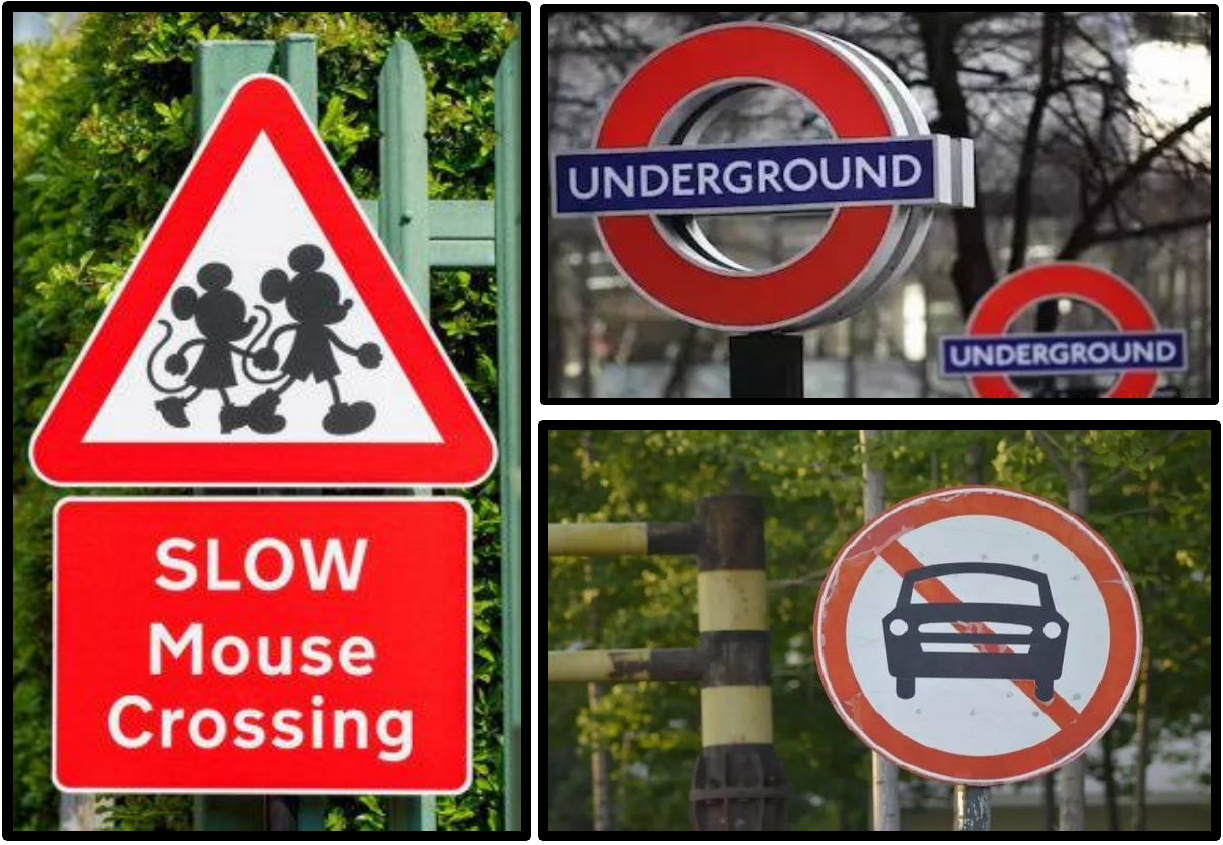}
    \label{fig:firstpagea}
    }
    \subfloat[]{
    \includegraphics[width=0.45\linewidth]{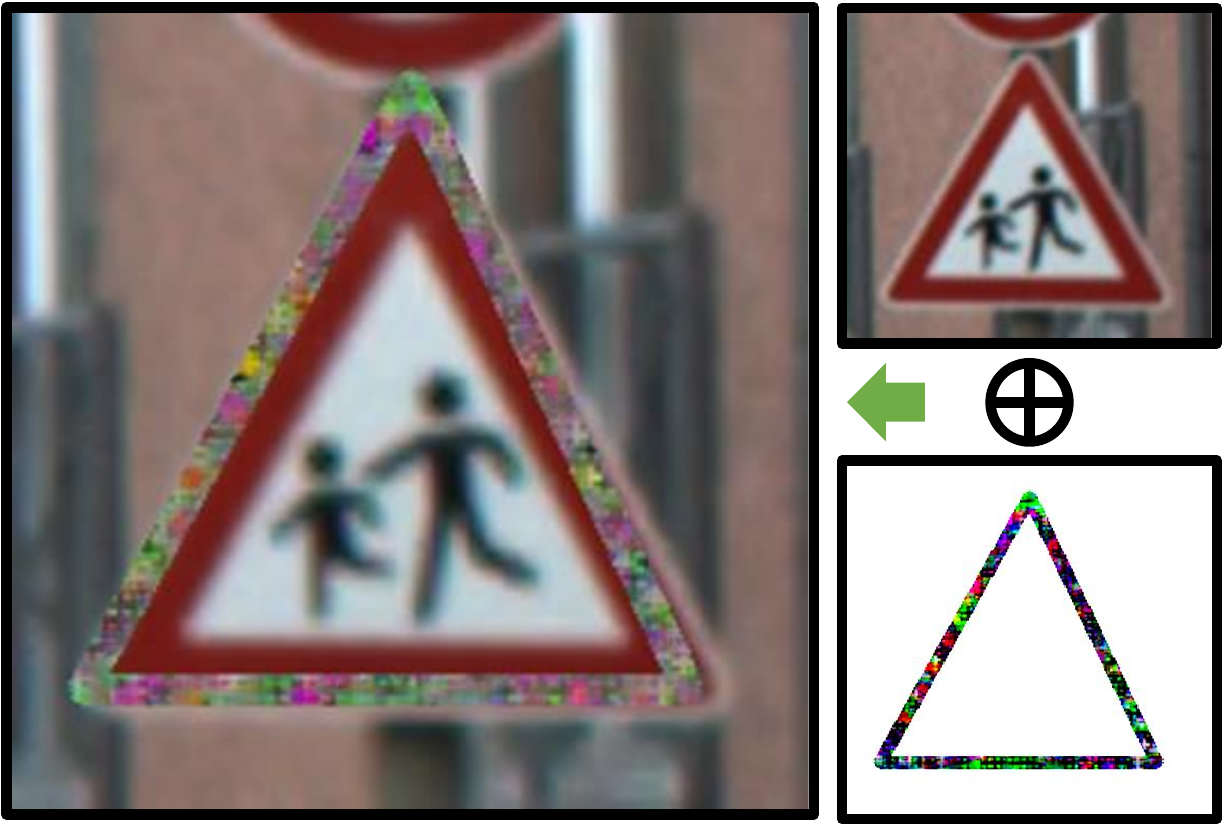}
    \label{fig:firstpageb}
    }\\
    \subfloat[]{
    \includegraphics[width=0.45\linewidth]{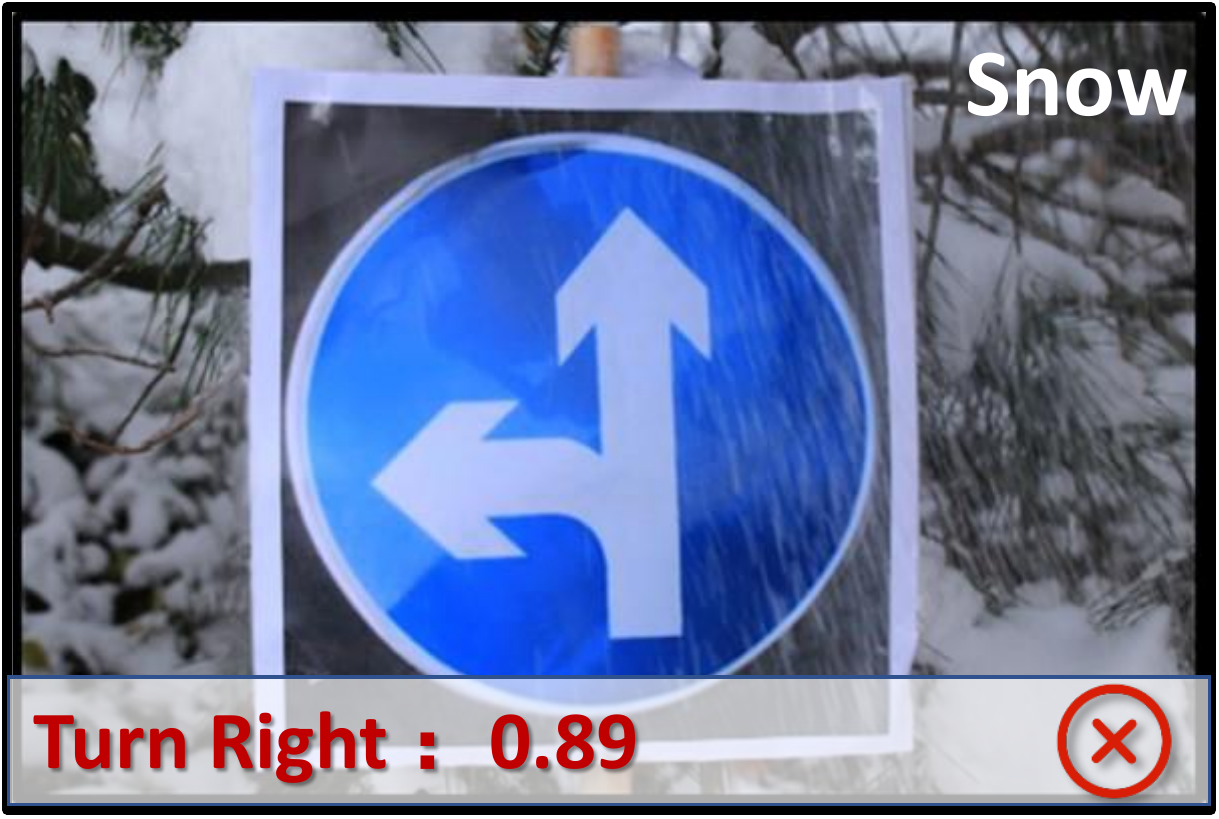}
    \label{fig:firstpagec}
    }
    \subfloat[]{
    \includegraphics[width=0.45\linewidth]{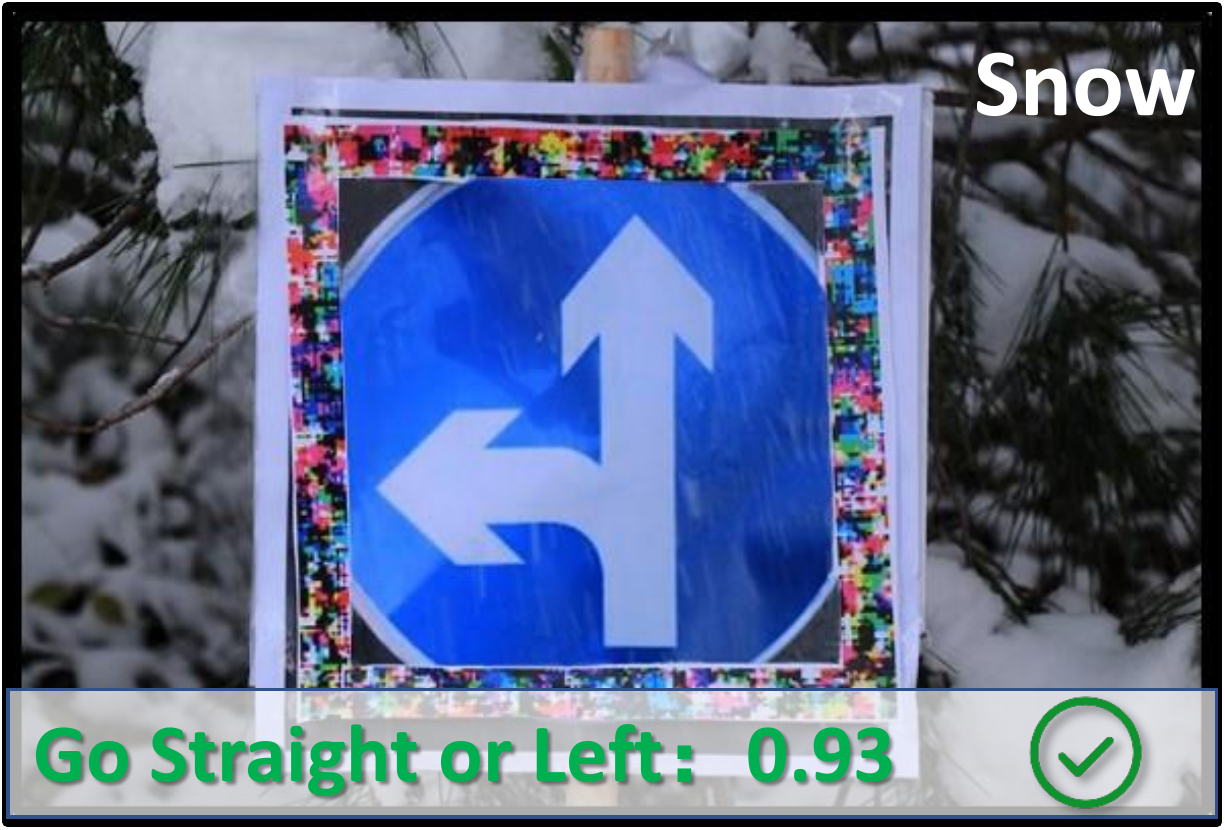}
    \label{fig:firstpaged}
    }
    \caption{(a) Different guideboards in the physical world. (b) Samples with generated defensive patches in the digital world. (c) The model prediction is misled into \texttt{Turn Right} when it is snowy in the physical world. (d) Defensive patches can help models to conduct correct predictions during snow in the physical world.}
    \label{fig:my_label}
    \vspace{-0.15in}
\end{figure}

Though deep neural networks (DNNs) have achieved significant successes in multiple areas \cite{DCTNet,li2021generalized,yin2021improving}, their robustness is challenged by noises, especially in physical world scenarios. Adversarial noise, an imperceptible perturbation designed to mislead the decision of DNNs, is now becoming a great threats \cite{szegedy2013intriguing,goodfellow2014explaining}. 
Besides adversarial attacks, DNNs also show weak robustness against common corruptions in the daily environment (\eg, snow, rain, brightness \etc) \cite{hendrycks2019benchmarking,hendrycks2020many}.
For example, the guide boards will be incorrectly classified as \texttt{Turn Right} when it is snowy (Figure (\ref{fig:firstpagec})). What's worse, these inevitable noises have caused dozens of self-driving accidents with casualties and are casting a shadow over the deep learning applications in practice \cite{avcrash}. This urges us to investigate feasible defenses for building robust deep learning models in the physical world.

In the past years, a great number of efforts have been made to defend against the adversarial perturbations and further improve model robustness \cite{madry2017towards,yuan2018commandersong,samangouei2018defense,hao2015sparse,liu2021ANP}. Most of the existing works focus on enhancing robustness from model-end (\eg, data augmentation, adversarial training), which require an additional cost of the model architecture modification or model retraining. In contrast, another line of studies performs defenses from the data-end without imposing any model modification (\eg, input transformation), which has shown great potential in practice \cite{DBLP:journals/corr/abs-1711-01991,salman2020unadversarial}. For example, by simply sticking a patch on the traffic sign, our proposed defensive patch can help DNNs to make robust recognition under noises (Figure \ref{fig:firstpaged}). Though showing great application potential, existing data-end defenses show several limitations when applied in practice: (1) Weak generalization for diverse noises. Existing works show a significant drop when facing different unseen noises (\eg, adversarial attacks, common corruptions). (2) Low transferability across multiple models. In other words, these works fail to perform defenses for black-box models and even being counteractive. We attribute this phenomenon to the underutilization of robust recognition characteristics.

To address the problems mentioned above, this paper proposes a data-end defensive patch generation framework, which could be effective against diverse noises and work among different models (Figure (\ref{fig:firstpageb}) and Figure (\ref{fig:firstpaged})).
Previous studies have revealed strong evidence that robust recognition highly depends on the exploitation of local and global features \cite{9199576, XIONG2017225,pirgazi2021efficient}, we thereof improve the defense ability of our defensive patches by promoting better exploitation of both local and global features. 
Regarding the generalization against diverse noises, since deep learning models rely strongly on the local patterns for predictions \cite{lee2021robust,hernandez2021multi,lu2020comparative}, we optimize the locally confined patch priors to contain more class-specific identifiable patterns via reducing model uncertainty. Based on these class-level patch priors, the defensive patches can preserve more recognizable features for a specific class and help models to better resist the influence of different noises, \ie, better generalization. As for the transferability across multiple models, recent studies found that different models share similar global perception during decision-making \cite{kim2019neural, Wang_2021_CVPR,curby2019behind}, we thus guide the defensive patches to capture more class-wise global feature correlations. In other words, the defensive patches could contain more global features correlated to the class. Thus, the generated defensive patches could better activate the model-shared global perception and enjoy stronger transferability among multiple models. 
In conclusion, our main contributions can be summarized as:
\begin{itemize}
\setlength{\itemsep}{-0.5ex}
    \item To the best of our knowledge, we are the first to generate data-end defensive patches that could improve application robustness against diverse noises (adversarial attacks and corruptions) among different models.
    \item Our defensive patches improve robustness by injecting local identifiable patterns and enhancing global perceptual correlations, which can be easily deployed via sticking them around target objects.
    \item Extensive experiments show that our defensive patch outperforms others by large margins (\textbf{+20\%} accuracy for both adversarial and corruption robustness on average in digital and physical world).
\end{itemize}

\section{Related Work}
\label{sec:related}

\subsection{Adversarial Attacks}

Extensive studies have shown that deep learning models are highly vulnerable to adversarial attack \cite{szegedy2013intriguing,goodfellow2014explaining,wei2019adversarial,Liu2020Spatiotemporal}. These imperceptible perturbations could easily make DNNs misclassify the input images. Besides adversarial perturbations, adversarial patches are designed to attack DNNs by attaching additional stickers for their feasibility in the physical world \cite{brown2017adversarial,liu2020bias,liu2019perceptual,Wang_2021_CVPR,wang2021Aco,Duan_2020_CVPR}, including patches \cite{brown2017adversarial}, camouflages \cite{Wang_2021_CVPR}, and light \cite{duan2021adversarial}. \cite{brown2017adversarial} proposes the first adversarial patch generation strategy, revealing the possibility of generating physical adversarial examples. \cite{Wang_2021_CVPR} generates patch-like adversarial camouflage in a 3D environment by suppressing model and human attention. Some researchers aim to perform adversarial attacks in the physical world with adversarial lights (\eg, laser \cite{duan2021adversarial} and infrared light \cite{zhu2021fooling}). 

Besides adversarial attacks, there exist another type of noise named common corruptions, which are commonly-witnessed natural noises, \eg, blur, snow, and frost, \etc. A line of works has been devoted to studying the influence of common corruptions for DNNs by various approaches \cite{hendrycks2019benchmarking,hendrycks2021natural,hendrycks2020many,tang2021robustart}. \cite{hendrycks2019benchmarking} proposes a challenging datasets on ImageNets (\ie, ImageNet-C), which contain 15 different types of common corruptions.  \cite{hendrycks2021natural} find that unmodified examples can mislead various unseen models reliably. In summary, the robustness of DNNs is highly challenged by the diverse noises in the physical world, which urges us to improve the 
application robustness and applicability.

\subsection{Adversarial Defenses}

Adversarial defenses aim to improve the robustness against adversarial attacks, which play important roles in increasing the availability of DNNs in the real world. Recent studies indicate that there exists three mainstreams in adversarial defenses: (1) Gradient masking, which aims to hide the key information of the model (\ie, gradients), including defensive distillation \cite{papernot2016distillation}, shattered gradients \cite{guo2017countering}, randomized gradient \cite{dhillon2018stochastic}, \etc. (2) Adversarial training, which improves the model robustness through adversarially training the classifier with adversarial examples \cite{goodfellow2014explaining,madry2017towards,tramer2017ensemble,shafahi2019adversarial,wong2020fast,liu2021ANP,xie2020smooth}. (3) Adversarial example detection, which aims to distinguish whether the input is clean or adversarial example \cite{grosse2017statistical,gong2017adversarial,jiang2020attack}. The above-mentioned studies primarily focus on improving 
application robustness from the model-end, which requires the model architecture modification or model re-training. Besides, there also exists another type of defense, which is feasible in the real world by modifying the input data (\ie, data-end defenses), such as input transformation \cite{xie2017mitigating} or image compression \cite{Jia_2019_CVPR}.

In this paper, we focus on the data-end defense and design a defensive patch to improve application robustness against diverse noises in the physical world.

\section{Approach}
\label{sec:approach}
In this section, we first give the definition of the defensive patch and then elaborate on the proposed framework.
\subsection{Problem Definition}
A perturbed example $x'$, which consists of a clean example and additional noises, can mislead a given deep neural network $\mathbb{F}$ into wrong prediction, \ie, $\mathbb{F}(x) \neq \mathbb{F}(x')$. Given a $k$-class dataset $\mathcal{X}=\mathcal{X}^1\cup\mathcal{X}^2\cup\cdots\cup\mathcal{X}^k$, the clean example $x \in \mathcal{X}$ and its corresponding perturbed example $x'$ are subject to a $\epsilon$-constraint. Base on the above knowledge, we now provide the definition of defensive patch $\delta$ as
\begin{equation}
\mathbb{F}(x) = \mathbb{F}(x'\oplus\delta)\quad s.t. \quad \lVert\delta\rVert \leq \varepsilon,
\end{equation}
where $\lVert\cdot\rVert$ is a distance metric which is often measured by $\ell_p$-norm ($p\in$\{1, 2, $\infty$\}), and $\varepsilon$ is a constraint value. The defensive patch $\delta$ also satisfies the $\mathbb{F}(x) = \mathbb{F}(x\oplus\delta)$ constraint. And the operation $\oplus$ obeys the following equation
\begin{equation}
\label{equ:mask}
    x \oplus \delta = (\textbf{1}- \operatorname{\textbf{M}})\odot x + \operatorname{\textbf{M}} \odot \delta,
\end{equation}
where $\odot$ is the element-wise multiplication and $\operatorname{\textbf{M}}$ is a shape mask to decide the masking position and appearance.

\begin{figure*}[!]
\begin{center}
\includegraphics[width=0.95\linewidth]{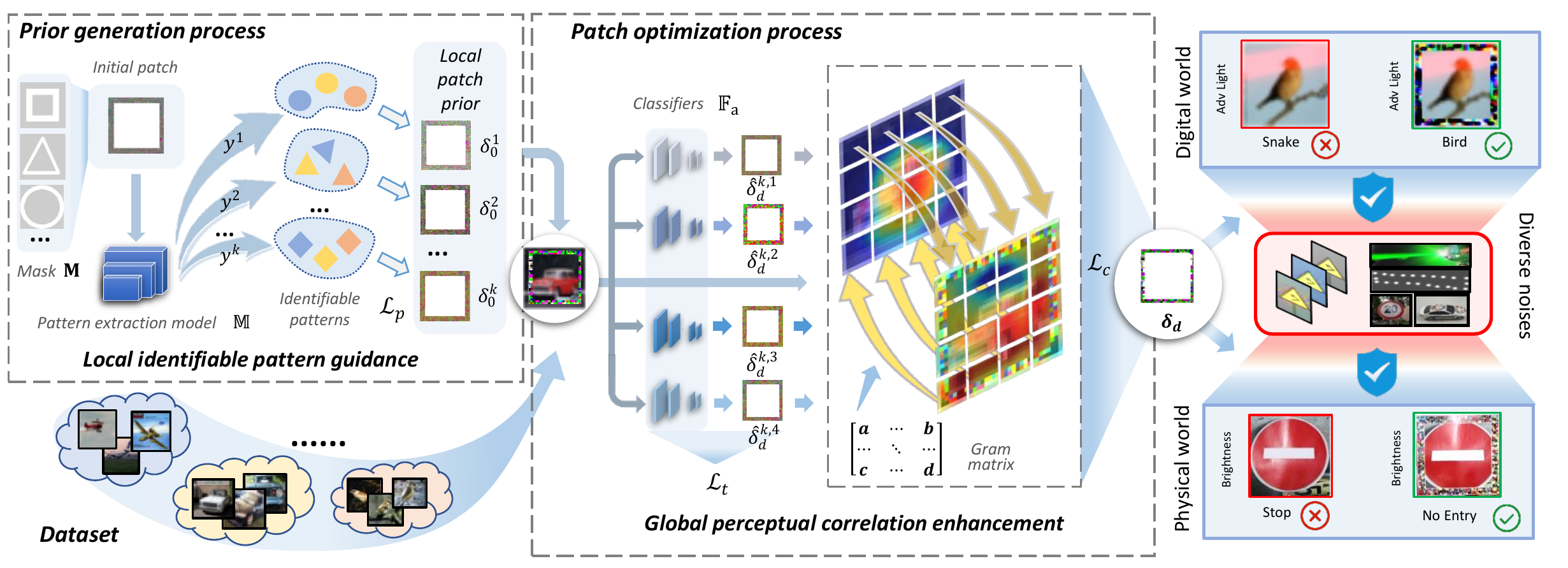}
\end{center}
\caption{Defensive patch generation framework. We first generate a typical patch prior for each category and inject more class-specific identifiable patterns into the confined local patch from the local viewpoint. Further, we help the defensive patches to contain more global features correlated to each category in an ensemble way from the global viewpoint. Therefore, the generated defensive patches enjoy both strong generalization and transferability.}
\label{fig:framework}
\vspace{-0.15in}
\end{figure*}

\subsection{Framework Overview}

Previous studies have indicated that robust recognition shows high dependence on the combination of local and global features \cite{9199576, XIONG2017225,pirgazi2021efficient}, we thus propose to generate defensive patches with strong noise generalization and model transferability by helping models for the better exploitation of local and global features. Thus, our defensive patches can significantly improve the robustness of recognition. The overall framework is shown in Figure \ref{fig:framework}.

Regarding the generalization against diverse noises, inspired by the fact that deep learning models recognition depends heavily on local patterns \cite{lee2021robust,hernandez2021multi,lu2020comparative}, we inject more class-specific identifiable patterns into the confined local patch prior. Thus, the defensive patch optimized from the patch prior could preserve more class-specific recognizable features, which could lead the model to better recognition under diverse noises. As for the transferability across multiple models, since different models focus on the similar global perception when making decisions \cite{kim2019neural, Wang_2021_CVPR,curby2019behind}, we guide the defensive patches to capture more global feature correlations within a class using Gram matrix in an ensemble way. Thus, our defensive patches could better activate model-shared global perceptions and show stronger transferability among models.

\subsection{Local Identifiable Patterns Guidance}

Several previous studies have pointed out that deep learning models show a strong dependence on local patterns \cite{lee2021robust,hernandez2021multi,lu2020comparative}, \eg, local patterns are exploited to improve the emotion recognition ability of the model \cite{hernandez2021multi}. Therefore, we aim to inject more class-specific identifiable patterns into the confined local patch prior. The defensive patch optimized from the prior can be treated as a typical class-specific representation \cite{liu2020bias}, hence helping the model for better recognition under different noises.

In practice, we first consider the shape of the local patch prior. Since the defensive patch is designed to improve application robustness in the real-world scenario, it is necessary to evade the influence for human vision (\ie, without covering the target object). Thus, we set the shape mask $\operatorname{\textbf{M}}$ in Equation \ref{equ:mask} as a $w$-pixels square box surrounding the target object (\ie, like guideboard border). Thus, the initial patch prior $\delta$ is reformulated as
\begin{equation}
    \delta = \textbf{0} + \operatorname{\textbf{M}}\odot \textbf{1},
\end{equation}
where the $\mathbf{0}$ and $\mathbf{1}$ are respectively a tensor in which each element is 0 or 1, and their dimensions are the same with input size of $\operatorname{\textbf{M}}$.
Note that the position mask can be replaced with any different shapes based on the scenario (see studies in Section \ref{sec:shapeablation}).

To inject more class-specific identifiable patterns into the confined local prior, we borrow a pattern extraction model to optimize the patch prior by an entropy-based loss function. Since the entropy is widely used to depict the class uncertainty, \ie, higher entropy indicates higher uncertainty to recognize the object. We thus force the patch priors to reduce the entropy of a certain class, \ie, making it more recognizable for a specific class. In this way, the defensive patches can be optimized to contain more class-specific identifiable patterns and resist the influence of different noises. In particular, given a pattern extraction model $\mathbb{M}$ and specific class index $k$, we optimize the $\delta_0^k$ (initialized as $\delta$) by calculating the identifiable pattern loss $\mathcal{L}_p$ as
\begin{equation}
    \mathcal{L}_p = -\log{\mathcal{P}_{\mathbb{M}}(\delta_0^k}),
\end{equation}
where $\mathcal{P}_{\mathbb{M}}$ is the prediction value of $\mathbb{M}$ with class index $k$. It is important to note that no input data is needed in the patch prior generation process and $\mathbb{M}$ could be any pre-trained model for this task.

Due to the fact that the typical patch priors contain more identifiable patterns, the defensive patches optimized from these priors could preserve more recognizable features towards a specific class and show better generalization against different noises. After the patch prior generation, we exploit the typical patch prior $\delta_{0}^k$ in the following defensive patch optimization procedure.

\subsection{Global Perceptual Correlation Enhancement}

Motivated by the fact that models often share similar global perception when making correct predictions towards a specific class \cite{kim2019neural,Wang_2021_CVPR,curby2019behind}, we aim to improve the transferability among different models by better activating the model global perception.

Since the context of the target objects is essential as well for making a correct perception \cite{goferman2011context}, we make the defensive patches capture more globally contextual features within a certain class. Considering the fact that Gram matrix can be exploited to represent the correlations of features within an image \cite{2016Image}, we design a global correlation loss based on Gram matrix by introducing stronger global feature correlations with respect to the class in an ensemble approach. Therefore, the generated defensive patches can better activate the model global perception and achieve stronger transferability among different models.

Specifically, given classifiers $\mathbb{F}_i$ ($i = 1, 2, ..., \operatorname{N}$), we first initialize the defensive patch $\delta_d^{k}$ of the $i$-th class as $\delta_0^{k}$; we then optimize the intermediate defensive patch of the $i$-th classifier $\hat{\delta}_{d}^{k,i}$ from the $\delta_d^{k}$  using clean examples $x \in \mathcal{X}^k$ based on $\mathcal{L}_t$ as
\begin{equation}
    \mathcal{L}_t = y^k\cdot\log \mathcal{P}_{\mathbb{F}_i}(x\oplus \hat\delta_{d}^{k,i}),
\end{equation}
where $y^{k}$ denotes the ground-truth label of $x$, $\mathcal{P}_{\mathbb{F}_i}$ denotes the prediction value of $\mathbb{F}_i$ with the input $x\oplus \hat\delta_{d}^{k,i}$.  

We then optimize our defensive patches by exploiting the most similar global perception shared among these intermediate patches from different models. In detail, we introduce the Gram matrix to optimize $\delta_d^k$ based on the combination of multiple $\hat{\delta}_{d}^{k,i}$ by perceptual correlation loss $\mathcal{L}_c$ as

\begin{equation}
\label{equ:correlation}
\begin{gathered}
    \mathcal{L}_c = \frac{1}{\operatorname{N}}\sum_{i}^N\lVert \operatorname{\textbf{G}}(x\oplus \delta_d^k)-\operatorname{\textbf{G}}(x\oplus \hat{\delta}_{d}^{k,i})\rVert^2_2, \\
    \operatorname{\textbf{G}}_{p,q}(\mathbf{I}) = \sum_c \mathbf{I}_{pc}\cdot \mathbf{I}_{qc},
\end{gathered}
\end{equation}
where $\operatorname{\textbf{G}}_{p,q}(\mathbf{I})$ means the Gram matrix value of input $\mathbf{I}$ at position $(p, q)$, and $\mathbf{I}_{\cdot c}$ indicates the pixel value of the input $\mathbf{I}$ at channel $c$. We conduct the above defensive patch generation process in a progressive manner, and the optimized defensive patch during each iteration will serve as the prior for the next iteration.

Besides, it should be noted that this optimization process could also work under the single model setting, \ie, $\operatorname{N}$=1 (See experiments in Section \ref{sec:digitalsec}). We hypothesize the reason might be that as the high-order interaction, the global perceptual correlation perceived by a model plays an important role in robust recognition against attacks \cite{ren2021unified}.

To sum up, through enhancing the global perceptual correlations in an ensemble approach, the generated defensive patches can enjoy stronger transferability across multiple models by activating model-shared global perception. 

\subsection{Overall Training Process}
We generate the defensive patch by serially conducting two optimization processes, \ie, generating the typical patch prior by the identifiable pattern loss $\mathcal{L}_p$ and optimizing the defensive patch by the training loss $\mathcal{L}_t$ and the perceptual correlation loss $\mathcal{L}_c$.

Specifically, for each class, we first initialize all typical patch priors as $\delta$. Then we optimize the patch prior of the $k-$th class by minimizing $\mathcal{L}_p$ with a pattern extraction model $\mathbb{M}$ and the local position constraint $\operatorname{\textbf{M}}$.  
Furthermore, we employ $\operatorname{N}$ different models to conduct an ensemble-based perceptual correlation enhancement optimization. In detail, for each epoch, we obtain $\operatorname{N}$ intermediate defensive patches $\hat{\delta}_{d}^{k,i}$ by maximizing the training loss $\mathcal{L}_t$. After that, we minimize $\mathcal{L}_c$ to generate the defensive patches $\delta_d^k$ and then perform defenses by simply using them as additional ornaments.
Note that we set the $\operatorname{N}$ as 4 in this paper.

\section{Experiments}
\label{sec:expe}
In this section, we first illustrate our experimental settings, then evaluate the effectiveness of our defensive patch in both the digital and physical world.
\subsection{Experimental Settings}

\textbf{Datasets and models.} For the dataset, we choose the widely-used CIFAR-10 \cite{krizhevsky2009learning} and GTSRB (guideboard classification dataset) \cite{stallkamp2011german}. Regarding the models, we select the commonly used architectures including VGG-16 (denote ``VGG'') \cite{vgg19}, ResNet-50 (denote ``RNet'') \cite{ResNet50}, ShuffleNet-V2 (denote ``SNet'') \cite{ma2018shufflenet}, and MobileNet-V3 (denote ``MNet'') \cite{howard2019searching}. 

\textbf{Diverse noises.} In this paper, we employ 3 types of noises which are realizable in the physical world, \eg, corruptions \cite{hendrycks2019benchmarking}, AdvP \cite{brown2017adversarial}, and AdvL \cite{duan2021adversarial}. Specifically, for corruptions, we adopt the strategies from \cite{hendrycks2019benchmarking} and implement 16 kinds of corruptions such as fog, rainy, Gaussian, and light, \etc. For each corruption, we select 5 different intensities.

\textbf{Evaluation metrics and compared baselines.} To evaluate the performance of our proposed method, we choose the widely used metric $accuracy$ as the evaluation metric (the higher the better) following \cite{salman2020unadversarial}. As for the compared baselines, we employ UnAdv \cite{salman2020unadversarial} and Trans \cite{xie2017mitigating}, which are the state-of-the-art data-end defenses. We use their released codes for implementation and select reasonable settings for fair comparisons.

\textbf{Implementation details.}
For the hyper-parameter $a$, we set it as 4, which means 4 different models are employed. For the shape mask $\operatorname{\textbf{M}}$, we design a $w$-pixels bold box surrounding the object which constrains the patch size to 1/5 of image size following one of the implementations in \cite{salman2020unadversarial}. 
The backbone of the pattern extraction model $\mathbb{M}$ is VGG-19 \cite{vgg19}. During the prior and patch generation process, we use Adam optimizer with the learning rate of 0.01, weight decay of ${10}^{-4}$, and a maximum of 20 epochs. All codes are implemented in PyTorch. We conduct the training and testing processes on an NVIDIA GeForce RTX 2080Ti GPU cluster\textsuperscript{\ref{note:supp}}.

\subsection{Digital World Evaluation}
\label{sec:digitalsec}
In this section, we first evaluate the performance of our generated defensive patches in the digital world. Note that we select the public dataset CIFAR-10 to conduct digital world experiments.

Since our defensive patch generation framework employs several different models, it is unfair to directly compare our method with other baselines. Therefore, we conduct 2 different experiments respectively: (1) we generate our defensive patch by only using the same single target model with UnAdv; (2) we perform similar ensemble training for both UnAdv. Since Trans performs defenses without requiring target models, we directly report its results\textsuperscript{\ref{note:supp}}.


According to Table \ref{tab:onedigitalcifar} and Table \ref{tab:fourdigitalcifar}, we can conclude that our defensive patches show better performance for improving model application robustness, \ie, generalization against diverse noises and transferability among different models. We provide several conclusions as follows:

\begin{table*}[t]
\resizebox{\textwidth}{20mm}{
\begin{tabular}{cc|cccc|cccc|cccc|cccc}
\hline
\multicolumn{1}{c}{\multirow{2}{*}{Models}} & \multirow{2}{*}{Methods} 
            & \multicolumn{4}{c|}{VGG}  & \multicolumn{4}{c|}{RNet}  & \multicolumn{4}{c|}{SNet}   & \multicolumn{4}{c}{MNet} \\
\cline{3-18} 
      &     & Raw & Cor & AdvL & AdvP & Raw & Cor & AdvL & AdvP  & Raw & Cor & AdvL & AdvP & Raw & Cor & AdvL & AdvP  \\ 
\hline
 \multicolumn{2}{c|}{Clean}    & 92.67 & 54.67 & 80.40 & 19.99 & 94.65 & 51.51 & 85.65 & 53.51 & 92.33 & 49.71 & 78.90 & 43.76 & 93.65 & 52.04 & 84.13 & 46.76 \\ \hline 
\multirow{2}{*}{VGG}       
       
        & UnAdv   &  99.60   &  94.93  &   98.64  &  55.51   &  81.07   & 37.15  &  68.54   &   \textbf{27.47}  &  66.32   &   35.29  &  55.90   &  20.50  &  69.58   &  35.54   &  59.06  &  18.95      \\ \cline{2-18} 
        & \textbf{Ours}    & \textbf{99.98} & \textbf{98.16}    & \textbf{99.87}    & \textbf{75.41}   & \textbf{84.03}     & \textbf{39.08}    &    \textbf{71.08}  &  27.12    & \textbf{69.85}    & \textbf{40.08} & \textbf{58.94}     & \textbf{20.52}   &    \textbf{77.64} & \textbf{43.82}    & \textbf{68.59}    & \textbf{20.04}    \\ \hline
\multirow{2}{*}{RNet}    
        
        & UnAdv   &  64.21   & 37.05    &  57.50   &  15.31  &  99.23   &  78.83   & 96.94    &  74.41   &  77.35   &  42.28   &   65.64  &  25.61   &  67.92   &  32.69   &  57.60   &  17.36   \\ \cline{2-18}
        & \textbf{Ours}    & \textbf{72.44}    & \textbf{45.66}    &  \textbf{66.00}   & \textbf{18.39} & \textbf{99.93}    & \textbf{90.90}    & \textbf{99.50}    & \textbf{92.22}     & \textbf{83.37}    & \textbf{52.33} & \textbf{72.93}    & \textbf{29.92}    & \textbf{78.45}     & \textbf{44.00} & \textbf{68.52}     & \textbf{22.62}     \\ \hline
\multirow{2}{*}{SNet}
       
        & UnAdv   &    56.12   &  31.22  &  50.12  &  15.84   &  87.79    &  42.84   &  74.76   &   28.77  &  99.56   &  87.96   &  97.66   & 78.72    &   69.48  &  33.14  &  58.41   &  18.33   \\ \cline{2-18}
        & \textbf{Ours}    & \textbf{62.30}    & \textbf{39.34}   & \textbf{57.86}    & \textbf{21.28}    &  \textbf{89.57}   &  \textbf{47.95}   &  \textbf{78.14}   &  \textbf{35.16}   & \textbf{99.96}     & \textbf{93.06}     & \textbf{99.62}     & \textbf{89.70}     & \textbf{76.41}     & \textbf{41.43}     & \textbf{66.76}     & \textbf{23.58}     \\ \hline
\multirow{2}{*}{MNet} 
       
        & UnAdv   &   68.57  &  39.80 & 62.02 & 17.02 & 84.45   &  38.95   &  71.19  &  25.94   &   71.76  &  38.46   &  59.37   & \textbf{23.54}    &  99.93   &  90.26   &   99.50  &   80.25  \\ \cline{2-18}
        & \textbf{Ours}    & \textbf{71.16}    & \textbf{45.84 }  & \textbf{65.01} & \textbf{19.48} & \textbf{86.13}     & \textbf{43.32}     & \textbf{73.64}     & \textbf{25.95} & \textbf{74.51}     & \textbf{43.75}     & \textbf{63.48}     & 21.47    & \textbf{99.99}     & \textbf{93.94}     & \textbf{99.87}     & \textbf{93.04}     \\ \hline
\end{tabular}
}
\caption{The experimental results under single model setting. Note that we do not compare with Trans \cite{xie2017mitigating} in this situation. It can be observed that ``Ours'' shows better generalization and transferability. Higher accuracy values are in bold, \ie, better performance.}
\label{tab:onedigitalcifar}
\end{table*}

(1) For generalization against diverse noises, it can be observed that our defensive patches achieve higher accuracy under almost all noises. For example, for the single model setting on RNet, our method yields up to \textbf{10.81\%} improvement compared with UnAdv under white-box settings; for ensemble setting on MNet, we outperform Unadv and Trans up to \textbf{44.11\%} and \textbf{22.88\%}, respectively.


(2) For transferability among different models, it can be clearly illustrated from Table \ref{tab:onedigitalcifar} that our proposed defensive patch show higher accuracy values compared with Unadv under black-box settings. For example, our proposed method yields \textbf{10.51\%} improvement on average against corruptions on RNet.

(3) Besides, we can witness that UnAdv with ensemble strategy shows lower defending ability compared to the single setting. More precisely, ensemble strategies decrease the performance of white-box and increase that of black-box on UnAdv. For example, UnAdv show 70.36\% on VGG and 62.55\% on RNet against corruptions in the ensemble setting, while the accuracy on VGG against corruptions is 98.16\% under the single model setting. We attribute this observation to the deficiencies of the average ensemble strategy, \ie, ignoring the exploitation of shared high-level characteristics such as correlation among global features.  

To sum up, our defensive patch generation framework achieves high generalization and transferability in practical performance, showing significant accuracy improvements under diverse noises among multiple models, \ie, \textbf{20.18\%} improvement on average for adversarial robustness and \textbf{31.10\%} improvement on average for corruption robustness on the mentioned 4 models.

\begin{table}[b]
\centering
\footnotesize    
\begin{tabular}{cc cccc}
\hline
\multicolumn{1}{c}{Noises} & Methods & VGG & RNet & SNet & MNet \\ \hline
\hline
\multirow{3}{*}{Raw}       
        & Vanilla   &  92.67  &  94.65  &  93.65  &  92.33    \\ \cline{2-6} 
        & UnAdv   &  88.57  &  95.51  &  82.00  &  73.14    \\ \cline{2-6} 
        & Trans   &  88.84  &  93.69  &  90.24  &  91.31    \\ \cline{2-6}
        & \textbf{Ours}    &  \textbf{99.27} &  \textbf{98.82}  &  \textbf{99.02}  &  \textbf{99.68}    \\ \hline
\multirow{3}{*}{Cor}    
        & Vanilla   &  54.67  &  51.51  &  52.04  &  49.71    \\ \cline{2-6}
        & UnAdv   &  70.36  &  62.55  &  54.48  &  36.38    \\ \cline{2-6} 
        & Trans   &  51.71  &  51.53  &  49.15  &  50.41    \\ \cline{2-6}
        & \textbf{Ours}    &  \textbf{91.02}  &  \textbf{76.04}  &  \textbf{83.26}  &  \textbf{87.37}    \\ \hline
\multirow{3}{*}{AdvL}
        & Vanilla   &  80.40  &  85.65  &  78.90  &  84.13    \\ \cline{2-6}
        & UnAdv   &  83.00  &  88.87  &  71.79  &  62.36    \\ \cline{2-6} 
        & Trans   &  72.71  &  81.61  &  73.53  &  78.65    \\ \cline{2-6}
        & \textbf{Ours}    &  \textbf{97.27}  &  \textbf{96.04}  &  \textbf{96.07}  &  \textbf{98.67}    \\ \hline
 \multirow{3}{*}{AdvP} 
        & Vanilla   &  19.99  &  53.51  &  43.76  &  46.76    \\ \cline{2-6}
        & UnAdv   &  29.24  &  49.39  &  31.71  &  19.80    \\ \cline{2-6} 
        & Trans   &  33.18  &  59.52  &  52.91  &  53.17    \\ \cline{2-6}
        & \textbf{Ours}    &  \textbf{40.83}  &  \textbf{70.81}  &  \textbf{67.67}  &  \textbf{64.82}    \\ \hline
\end{tabular}
\caption{The experimental results under four models ensemble setting on CIFAR-10 dataset. Unadv \cite{salman2020unadversarial} and Trans \cite{xie2017mitigating} show weak defense ability.}
\label{tab:fourdigitalcifar}
\end{table}

\subsection{Physical World Evaluation}

To evaluate the effectiveness in the physical world, we select the traffic sign classification task under the consideration of the popularity of autonomous driving and its huge potential for applications.  Therefore, we generate our defensive patches based on a widely-used traffic sign classification dataset, \ie, GTSRB, and then print them using an HP Color LaserJet Professional CP5225 printer.

We choose three different real-world traffic signs from the campus environment as shown in supplementary\textsuperscript{\ref{note:supp}}, \ie, \emph{speed-limited 20} (denote ``SL''), \emph{no entry} (denote ``NE''), and \emph{go straight or left} (denote ``GSL''). We choose 4 different situations (\eg, raw, snow, brightness, adversarial patch) to simulate the different noises in the real world. Further, for adversarial attacks, we employ the AdvPatch and stick them on the traffic sign. For each kind of traffic sign under each situation, we sample images from 3 distances (\ie, 0.5m, 0.75m, and 1m) and 3 orientations (\ie, front side, left side, and right side). Regarding the defenses, we use our defensive patches and UnAdv. Therefore, we obtain $12*9*3 = 324$ images as the physical world test set in total, including diverse noises (corruptions and adversarial examples). Furthermore, we evaluate these real-world sampled images by RNet models to validate the practical effectiveness of our proposed method.


According to Table \ref{tab:phusicalresult}, we can observe that under different situations in the real world, our proposed defensive patches achieve better performance and outperform others (\ie, Vanilla and UnAdv) by large margins, \ie, higher accuracy values.
For all noises, our method achieves \textbf{26.86\%} improvement on average\textsuperscript{\ref{note:supp}}. 

Besides the above task, it should be noted that the proposed defensive patch generation framework owns the potential to perform defenses in other approaches, \eg, product special clothes, camouflages, coating, \etc. We generate some simple examples to demonstrate this viewpoint\textsuperscript{\ref{note:supp}}.

\begin{table}[b]
\centering
\footnotesize
\begin{tabular}{clcccc}
\hline
\multicolumn{1}{c}{\multirow{2}{*}{Guide board}} & \multicolumn{1}{c}{\multirow{2}{*}{Class}} & \multicolumn{4}{c}{Accuracy} \\ \cline{3-6} 
& \multicolumn{1}{c}{}& Raw    & Snowy    & Brightness & AdvP   \\ 
\hline\hline
\multirow{3}{*}{SL}                        
& Vanilla      &  44.44 &  44.44   &  33.33 & 11.11  \\ \cline{2-6} 

& UnAdv     &  33.33 &  33.33   &  11.11 & 22.22   \\ \cline{2-6} 
& \textbf{Ours}      &  \textbf{55.56} &  \textbf{55.56}   &  \textbf{44.44} & \textbf{44.44}  \\
\hline 
\multirow{3}{*}{NE}                   
& Vanilla      &  88.89 &  77.78   &  88.89 & 44.44   \\ \cline{2-6} 
& UnAdv     &  77.78 &  88.89   &  88.89 & 33.33   \\ \cline{2-6} 
& \textbf{Ours}      &  \textbf{100.00} &  \textbf{100.00}    &  \textbf{100.00}  & \textbf{66.67}    \\
\hline 
\multirow{3}{*}{GSL}                   
& Vanilla      &  44.44 & 33.33    &  22.22 & 33.33 \\ \cline{2-6} 
& UnAdv     &  44.44 & 11.11    &  33.33  & 33.33 \\ \cline{2-6} 
& \textbf{Ours}      &  \textbf{55.56} & \textbf{66.67}    &  \textbf{66.67} & \textbf{77.78}   \\
\hline 
\end{tabular}
\caption{Physical world experimental results. All images are tested on ResNet models. ``Ours'' accuracy value is much higher.}
\label{tab:phusicalresult}
\end{table}

\subsection{Discussion and Analysis}
In this section, we first provide some discussions from the perspectives of model attention (\ie, qualitative analysis) and decision boundary (\ie, quantitative analysis) to better understand our defensive patches; then we show that our defensive patches can be employed with other model-end strategies to further improve robustness.

\subsubsection{Model Attention Analysis}
We first adopt Grad-CAM \cite{Selvaraju_2017_ICCV} to visualize the attention of models when making predictions towards the same images with or without defensive patches. 

\begin{figure}[t]
    \centering
    \includegraphics[width=0.95\linewidth]{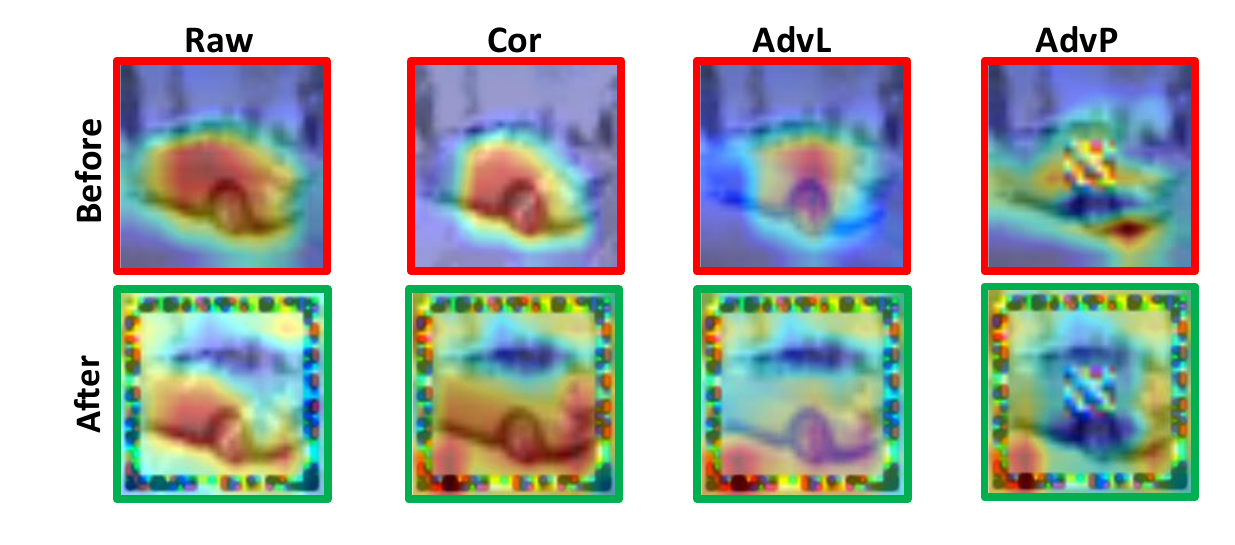}
    \caption{The model attention analysis. The red frames denote the examples without defensive patches (\ie, ``Before'') and the green ones denote the examples with defensive patches (\ie, ``After''). The model global perception is better activated. Best in view.}
    \label{fig:discussattention}
    \vspace{-.2in}
\end{figure}

Specifically, we select some samples from each category of CIFAR-10 and acquire their corresponding perturbed examples using noises (\ie, Cor, AdvL, AdvP). These instances satisfy the conditions that vanilla models fail to classify correctly on these perturbed images whereas they could provide correct predictions with the help of our defensive patches. Then we calculate the attention map of each group of the sampled images to exhibit the model perception variation by the Grad-CAM \cite{Selvaraju_2017_ICCV}. As shown in Figure (\ref{fig:discussattention}), after sticking the defensive patches, the model perception has been spread into a larger region globally over the image, which indicates that the model exploits more global features during the decision-making process\textsuperscript{\ref{note:supp}}. Thus, by better activating the global perception, our defensive patches could improve model application robustness and transfer among models.

\begin{figure}[b]
\vspace{-.2in}
    \centering
    \subfloat[]{
    \includegraphics[width=0.45\linewidth]{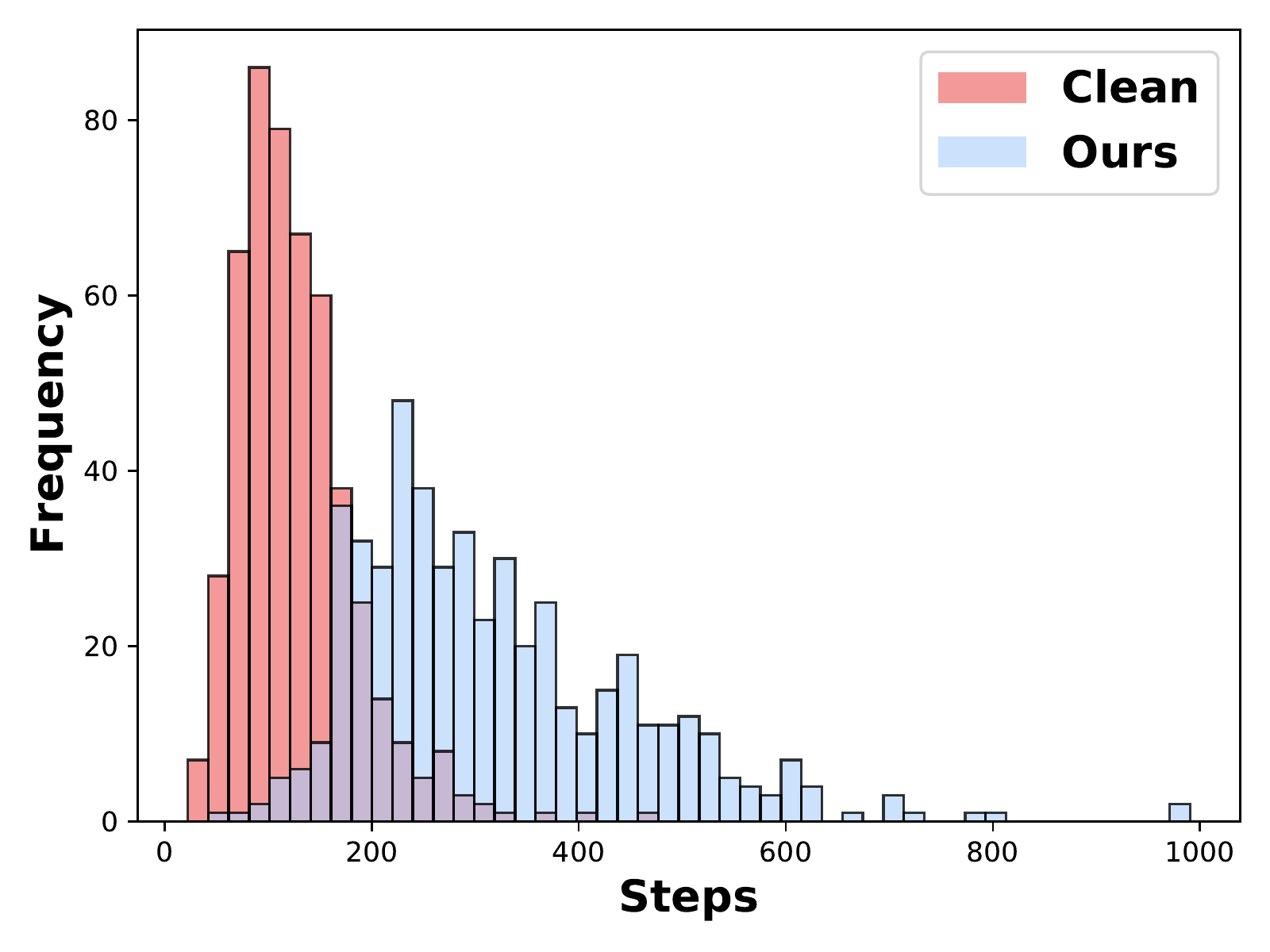}
    \label{fig:discussboundariesa}
    }
    \subfloat[]{
    \includegraphics[width=0.45\linewidth]{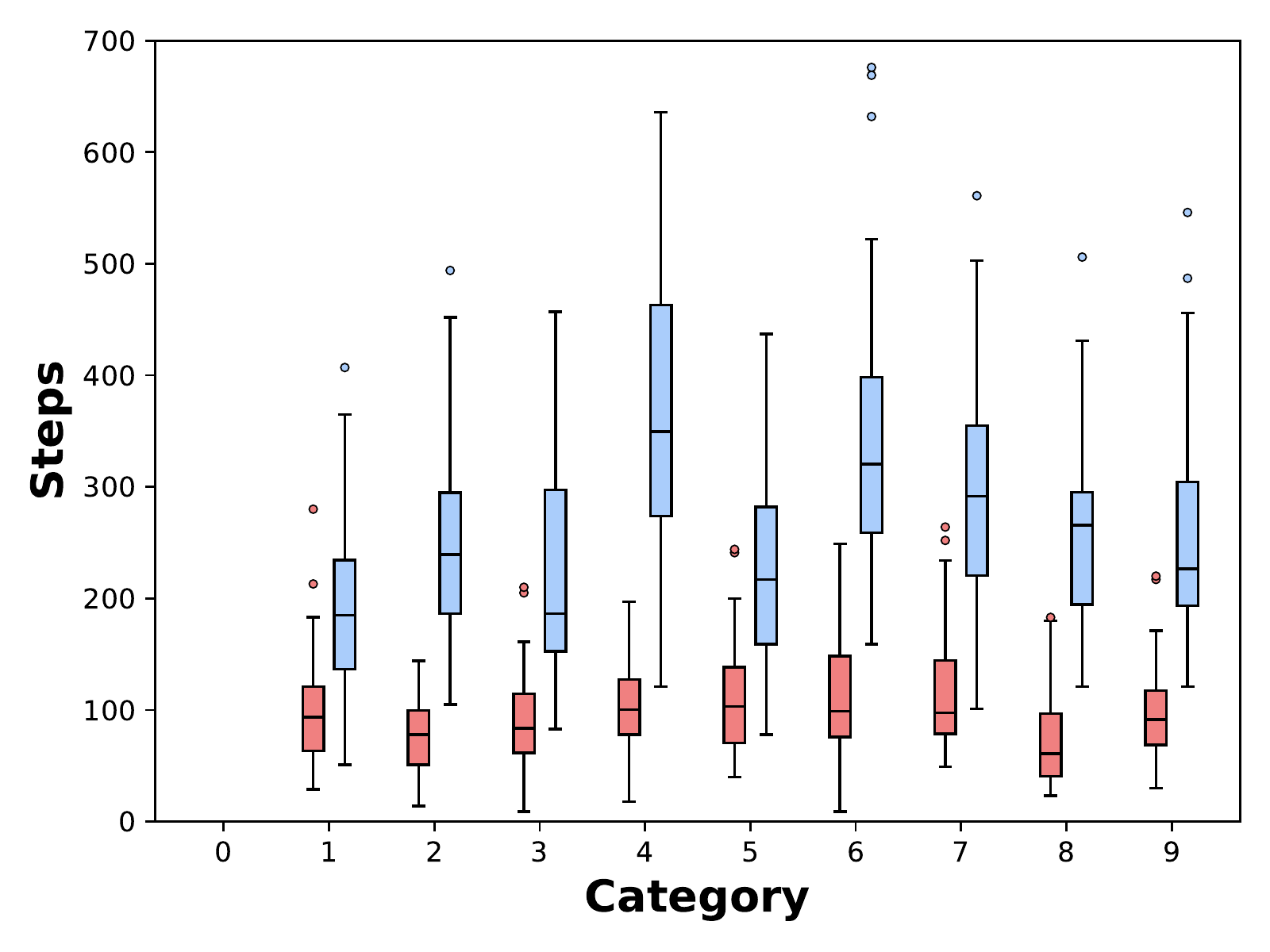}
    \label{fig:discussboundariesb}
    }
    \caption{Decision boundary analysis. (a) The frequency statistics for the number of average steps. (b) The distance statistics of a certain class, \ie, distances of class 0 to another 9 classes.}
    \label{fig:discussboundaries}
\end{figure}

\subsubsection{Decision Boundary Study}
To further understand our defensive patches, we follow \cite{liu2020bias} and provide a decision boundary analysis, which we aim to characterize the difficulty of fooling a classifier with or without our defensive patches. 

Specifically, we perturb an instance $x^i$ to specified classes and estimate the smallest optimization step numbers moved as the decision boundary distance. Given a learnt model $\mathbb{F}$ and a direction (\ie, class $y^j$, $i\neq j$), we optimize the instance until satisfying $\mathbb{F}(x^i) \neq y^i$ by following \cite{liu2020bias}. In detail, we randomly sample 50 examples for each category (500 in total) and employ RNet as $\mathbb{F}$. By calculating the statistics (\eg, average and median, \etc) of distances, we can draw some meaningful conclusions from Figure (\ref{fig:discussboundaries}). Firstly, it can be observed that the distribution of average step number shifts to bigger values after using our defensive patches (Figure \ref{fig:discussboundariesa}), which indicates that it is more difficult for adversaries to attack the models (\ie, the distance for different instances to decision boundaries are larger). Moreover, for each specific class, the decision boundary distances are larger after adding defensive patches as shown in Figure \ref{fig:discussboundariesb} (\eg, the blue boxes are higher than the red boxes). 

Therefore, we demonstrate that our defensive patches could help models to better resist the influence of noises, \ie, more difficult to be perturbed to other categories.  

\subsubsection{Combination with Other Model-end Defenses}
While the data-end defenses are independent to model-end defenses, it is rational for us to explore the possibility of jointly exploiting them both to further improve model application robustness against noises.

\begin{figure}[!]
    \centering
    \includegraphics[width=0.95\linewidth]{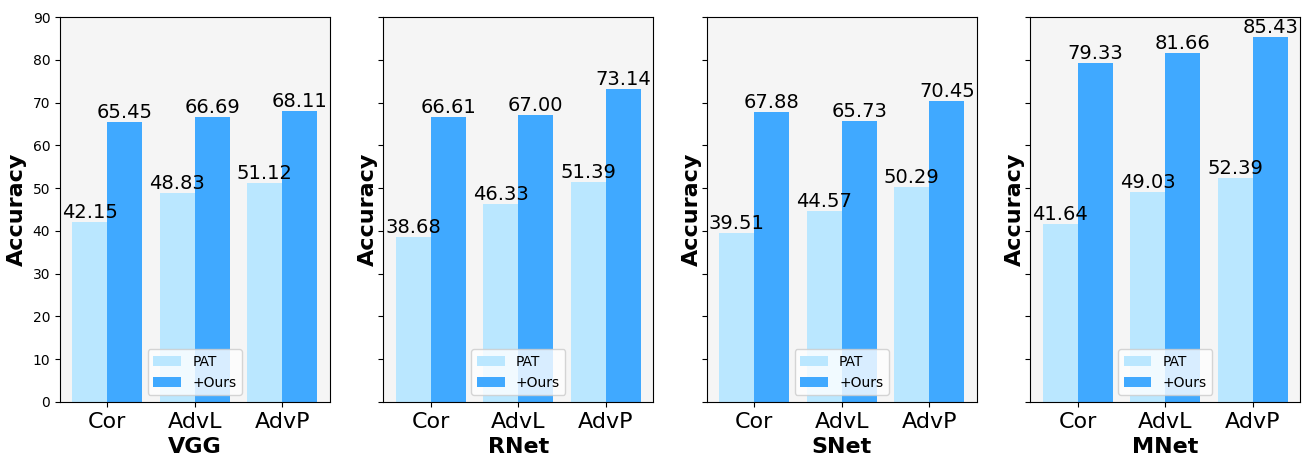}
    \vspace{-0.12in}
    \caption{The experimental results of working with model-end defenses. ``+Ours'' indicates the joint performance and achieves better performance, \ie, higher accuracy.}
    \label{fig:discussplug}
\end{figure}

In particular, we select the typical and popular model-end defense strategy, \ie, PAT \cite{2017Towards}, which adversarially train models with PGD adversarial attacks \cite{2017Towards}. We use VGG, RNet, SNet, and MNet as backbone models and adversarially train them respectively following the PAT strategies. For evaluation, we adopt the same testing dataset as Section \ref{sec:expe}. As illustrated in Figure \ref{fig:discussplug}, we can clearly observe the positive effects of the defensive patches, \ie, by adding our defensive patches with PAT, +\textbf{29.32\%} on corruptions, +\textbf{23.03\%} on adversarial noises. These results enable us to draw a meaningful conclusion that our defensive patches could serve as a strong method for real-world applications due to their flexible usage and significant promotion for robust recognition. Another model-end defense strategy, \ie, PatG \cite{2020PatchGuard}, is also compared\textsuperscript{\ref{note:supp}}.

\subsection{Ablations Studies}
In this section, we provide ablation studies to better understand the effectiveness of different parts of our defensive patch generation framework.
\subsubsection{Impact of Different Loss Terms}
Here, we first investigate the impacts of the different loss terms, \ie, $\mathcal{L}_{p}$ and $\mathcal{L}_{c}$.

Specifically, we first study the impact of the $\mathcal{L}_{p}$ to the generalization ability against diverse noises. To make it a fair comparison, we train two models with $\mathcal{L}_{t}$ and $\mathcal{L}_{t}$+$\mathcal{L}_{p}$ with the ensemble setting. According to Table \ref{tab:lossgeneralization}, we can observe that the robustness under ``$\mathcal{L}_{t}$+$\mathcal{L}_{p}$'' setting is much higher than that under $\mathcal{L}_{t}$ setting. The results empirically prove that $\mathcal{L}_{p}$ could improve model application robustness against diverse noises.

Then we study the impact of the $\mathcal{L}_{c}$ to the transferability across different models under the single model setting. In detail, we select the RNet as the source model and the other 3 models as target models (VGG, SNet, and MNet). As shown in Table \ref{tab:losstransfer-a}, \ref{tab:losstransfer-b}, and \ref{tab:losstransfer-c}, the ``$\mathcal{L}_{t}$+$\mathcal{L}_{c}$'' always achieves higher accuracy on almost all target models under different noise settings, which strongly support that $\mathcal{L}_{c}$ could improve the transferability between models\footnote{Please refer to the Supplementary Material for more details\label{note:supp}}.

\begin{table}[t]
\footnotesize
\centering 
\begin{minipage}[]{0.49\columnwidth}
\setcaptionwidth{0.9\columnwidth}
\centering
\setlength{\tabcolsep}{0.8mm}{
\begin{tabular}{cccc}
\hline
\multicolumn{1}{c}{\multirow{2}{*}{Method}} & \multicolumn{3}{c}{Noises}\\
\cline{2-4} 
\multicolumn{1}{c}{}   & Cor & AdvL & AdvP \\ \hline\hline
$\mathcal{L}_{t}$      &  61.73   &    88.90    &      46.45\\ \hline
$+\mathcal{L}_{p}$ & \textbf{71.21} & \textbf{94.35} & \textbf{59.02}\\ \hline
Ours       & \textbf{76.04}  & \textbf{96.04}&\textbf{68.81}\\ \hline
\end{tabular}}
\vspace{-0.1in}
\caption{Ablations on $\mathcal{L}_{p}$ for ``Cor'', ``AdvL'', ``AvP'' under ensemble settings.}
\label{tab:lossgeneralization}
\end{minipage}
\begin{minipage}[]{0.49\columnwidth}
\setcaptionwidth{0.9\columnwidth}
\centering
\setlength{\tabcolsep}{0.8mm}{
\begin{tabular}{cccc}
\hline
\multicolumn{1}{c}{\multirow{2}{*}{Method}} & \multicolumn{3}{c}{Models}\\ 
\cline{2-4} 
\multicolumn{1}{c}{}   & VGG & SNet & MNet \\ \hline\hline
$\mathcal{L}_{t}$      &  27.99 &  32.37  &  27.31\\ \hline
$+\mathcal{L}_{c}$     &  \textbf{34.51} & \textbf{43.03}& \textbf{30.19}\\ \hline
Ours       & \textbf{46.85}  & \textbf{50.35}&\textbf{45.86}\\ \hline
\end{tabular}}
\vspace{-0.1in}
\caption{Ablations on $\mathcal{L}_{c}$  under single model settings for ``Cor".}
\label{tab:losstransfer-a}
\end{minipage}
\begin{minipage}[!t]{0.49\columnwidth}
\setcaptionwidth{0.9\columnwidth}
\centering
\setlength{\tabcolsep}{0.6mm}{
\begin{tabular}{cccc}
\hline
\multicolumn{1}{c}{\multirow{2}{*}{Method}} & \multicolumn{3}{c}{Models}\\ 
\cline{2-4} 
\multicolumn{1}{c}{}   & VGG & SNet & MNet \\ \hline\hline
$\mathcal{L}_{t}$      &  49.56 &  57.21 & 51.98\\ \hline
$+\mathcal{L}_{c}$     &  \textbf{53.63} &\textbf{65.41}& \textbf{54.13}\\ \hline
Ours       & \textbf{66.80}  & \textbf{71.84}&\textbf{71.72}\\ \hline
\end{tabular}}
\vspace{-0.1in}
\caption{Ablation on $\mathcal{L}_{c}$ under single model settings for ``AdvL".}
\label{tab:losstransfer-b}
\end{minipage}
\begin{minipage}[!t]{0.49\columnwidth}
\setcaptionwidth{0.9\columnwidth}
\centering
\setlength{\tabcolsep}{0.8mm}{
\begin{tabular}{cccc}
\hline
\multicolumn{1}{c}{\multirow{2}{*}{Method}} & \multicolumn{3}{c}{Models}\\ 
\cline{2-4} 
\multicolumn{1}{c}{}   & VGG & SNet & MNet \\ \hline\hline
$\mathcal{L}_{t}$      & 13.60  & \textbf{25.62}  & 16.48\\ \hline
$+\mathcal{L}_{c}$     & \textbf{15.52} & 25.52 &  \textbf{16.60}\\ \hline
Ours       & \textbf{18.39}  & \textbf{29.92}&\textbf{22.62}\\ \hline
\end{tabular}}
\vspace{-0.1in}
\caption{Ablation on $\mathcal{L}_{c}$ under single model settings for ``AdvP".}
\label{tab:losstransfer-c}
\end{minipage}
\end{table}



\subsubsection{The Number of Ensemble Models}
Since our defensive patch generation framework introduces the ensemble strategy, it is necessary to investigate the effects on the number of ensemble models. 

Specifically, we adopt different ensemble settings (\ie, optimize the defensive patch based on 1, 2, 3, and 4 models), and keep other settings the same for fair comparisons. We can summarize the following observations: 
(1) application robustness improves with the increasing of ensemble model numbers;
(2) beyond the ``white-box'' models (\ie, the ensemble models), our generated patches perform better on unseen models.
Thus, we could conclude that the transferability between models is benefited from the ensemble perceptual correlation reinforcement\textsuperscript{\ref{note:supp}}.

\subsubsection{The Shape of the Defensive Patches}
\label{sec:shapeablation}
Finally, we investigate the performance of defensive patches with different shapes (\ie, different shape masks $\operatorname{\textbf{M}}$ in Equation (\ref{equ:mask}). Note that, this experiment is designed to test the defense ability of our patches in more practical scenarios.


Specifically, we handcraft 3 different shape masks, including circle, triangle, and trapezoid as shown in \emph{Section 3 in Supplementary Material}. Note that the sizes (pixel numbers) of these patches are set to be similar levels\textsuperscript{\ref{note:supp}}, which has no impact on the target object. We generate different defensive patches with different shape masks based on VGG, ResNet, ShuffleNet, and MobileNet, and then place them at the same positions and evaluate their performance (\ie, Raw, Cor, AdvL, AdvP). According to \emph{Table 7 in Supplementary Material}, we can conclude that the shape only has very limited impacts on the performance of the defensive patches, \ie, 99.78\%, 98.87\%, 99.93\% for circle, triangle, and trapezoid, respectively (Raw accuracy on VGG), which can be ignored in real applications\textsuperscript{\ref{note:supp}}. Therefore, the proposed defensive patch generation framework can be more flexible in real-world applications for various scenarios.

\section{Conclusion}
\label{sec:con}

This paper proposes a novel defensive patch generation framework to conduct data-end defense by better exploiting both local and global features. Our defensive patches could achieve strong generalization against diverse noises and transferability among different models. 
Extensive experiments demonstrate that our defensive patch outperforms others by large margins (\eg, improve \textbf{20+\%} accuracy for both adversarial and corruption robustness on average in the digital and physical world).

Our defensive patches could be easily deployed in practice to defend noises by simply sticking them around the target objects (\eg, traffic signs in cities that often snow). In the future, we are interested in trying more convenient approaches such as employing these patches as a pre-processing procedure (automatic detecting and sticking the defensive patch onto the image and then feeding the image to the system). Moreover, applying this strategy in more visual tasks, \eg, detection tasks, is also worth attempting.


\section{Acknowledgement}
This work was supported by the National Key Research and Development Plan of China under Grant 2020AAA0103502, the National Natural Science Foundation of China under Grant 62022009 and Grant 61872021, and the Beijing Nova Program of Science and Technology under Grant Z191100001119050.

\clearpage
{\small
\bibliographystyle{ieee_fullname}
\bibliography{egbib}
}

\end{document}